# WAVELET-FILTERING OF SYMBOLIC MUSIC REPRESENTATIONS FOR FOLK TUNE SEGMENTATION AND CLASSIFICATION

**Gissel Velarde**
Aalborg University
`gv@create.aau.dk`

**Tillman Weyde**
City University London
`T.E.Weyde@city.ac.uk`

**David Meredith**
Aalborg University
`dave@create.aau.dk`

**ABSTRACT**

The aim of this study is to evaluate a machine-learning method in which symbolic representations of folk songs are segmented and classified into tune families with Haar-wavelet filtering. The method is compared with previously proposed Gestalt-based method. Melodies are represented as discrete symbolic pitch-time signals. We apply the continuous wavelet transform (CWT) with the Haar wavelet at specific scales, obtaining filtered versions of melodies emphasizing their information at particular time-scales. We use the filtered signal for representation and segmentation, using the wavelet coefficients' local maxima to indicate local boundaries and classify segments by means of k-nearest neighbours based on standard vector-metrics (Euclidean, cityblock), and compare the results to a Gestalt-based segmentation method and metrics applied directly to the pitch signal. We found that the wavelet based segmentation and wavelet-filtering of the pitch signal lead to better classification accuracy in cross-validated evaluation when the time-scale and other parameters are optimized.

## 1. INTRODUCTION

One of the aims of folk song research is the study of melodic variations caused by the process of oral transmission between generations (van Kranenburg et al., 2009). Wiering et al. (2009) propose an interdisciplinary and ongoing process between human expertise, methods and models to understand melodic variation and its mechanisms. Classification models and methods dealing with such challenges define their representation and processing to be evaluated based on some ground truth. In this paper, we present our method based on wavelet-filtering and evaluate it on a collection of Dutch folk songs ("Onder de groene linde", Grijp, 2008), in which songs were classified into tune families according to expert similarity assessments, mainly based on rhythm, contour and motifs (Wiering et al., 2009; Volk & van Kranenburg, 2012).

The collection of folks songs that we study in this paper, is a monophonic collection of Dutch folk melodies encoded in MIDI files, so that we have pitches encoded as integer numbers, ranging from 0 to 127, and onsets and durations in quarter notes and subdivisions. In order to analyse these files via wavelets, we sample each melody as a one dimensional (1D) signal. Graphically, the melodic contour of 1D pitch signal can be drawn in a pitch over time plot, with the horizontal axis representing time in quarter notes, and the vertical axis representing pitch numbers. This contour representation of melodies has been linked to human melodic processing, using contour classes (Huron, 1996), interpolation lines (Steinbeck, 1982) and polynomial functions (Müllensiefen & Wiggins, 2011; Müllensiefen, Bonometti, Stewart & Wiggins, 2009). However, the contour representation does not give direct access to some aspects that are important for music similarity. Large-scale changes, like transposition of a melody lead to a completely different set of values although the melody is not substantially different. Similarly, small-scale changes like ornaments can lead to different pitch values even if the main essential shape of the melody is preserved.

Wavelet coefficients are obtained as the inner product of a 1D signal and a wavelet (i.e., a short signal with zero average and defined energy). The wavelet is shifted along the time axis and for each time position a coefficient is calculated. This is equivalent to a convolution with the wavelet flipped along the time axis, and thus to a finite impulse response filtering of the signal. The wavelet can be stretched on the time axis, leading to coefficients at different time-scales, corresponding to different filters. This process can also be understood as comparing the melodic shape with the wavelet shape, so that the coefficients represent similarity values at different time-positions and time-scales. The process of producing a full set of wavelet coefficients for a signal is known as the wavelet transform (WT), of which there are different variants. The transformed signal is represented as a set of coefficient signals at different scales. We use the Haar wavelet, which is a function of time $t$ that takes values of 1 if $0 \leq t < 0.5$, or $0.5 \leq t < 1$, and 0 otherwise.

We use the information of the wavelet coefficients to define and compare melodic segments. Local maxima of the wavelet coefficients occur when the inner product of the melody and the wavelet is maximal in that position. In the case of the Haar wavelet this occurs when there is a locally maximal change of pitch - averaged over half the length of the wavelet - in the melody. Therefore, we use the local maxima of wavelet coefficients to indicate segmentation points. If the found segments correlate with human structural perception and music theory, we assume that they can be used to classify melodies containing similar segments. A melodic fragment and its transposed version will be represented by the same wavelet coefficients (except for very beginning of the melody).

Musical similarity in folk music is a hard problem to define (Wiering et al., 2009). We can understand it as a partial identity, where entities share some properties that can be measured (Cambouropoulos, 2009). With wavelet-filtering we apply a process that selectively focuses on a specific time-scale. It is a preprocessing step before determining segment similarities, which we calculate based on distance metrics. In the following section we will discuss some computational models and methods that have been used to model melodic similarity in symbolic music representation and have been applied to classify folk melodies.

## 2. RELATED WORK

### 2.1 Modelling melodic variations

Computational models applied to modelling melodic variations in symbolic music representations of folk songs include string matching methods and multidimensional feature vectors to represent global properties of melodies (Hillewaere, Manderick & Conklin, 2009; Hillewaere, Manderick & Conklin, 2012; van Kranenburg, 2010). In origin and genre classification, global representations perform only slightly worse than string-based methods (Hillewaere et al., 2009 and 2012). However, methods based on global representation depend heavily on the choice of features, which can lead to reduce generalizability.

Van Kranenburg, Volk & Wiering (2013) showed that sequence alignment algorithms using local features prove successful in classifying folk song melodies to tune families defined by experts. Sequence alignment algorithms are used to quantify similarity of sequences by computing the operations needed to transform one sequence into another, by means of substitutions, insertions and deletions (Manderick & Conklin, 2012; van Kranenburg, 2010). Although van Kranenburg's (2010) method was very successful when used to classify melodies from the Dutch folk-song corpus into tune families, its representation requires 14 attributes for each note in a melodic sequence (see van Kranenburg, 2010, pp. 94-95), apart from the standard information that is encoded in MIDI format (pitch number, onset and duration), meaning that this approach might not be applicable for classification using MIDI files only. In the following section we present our method, which can be applied to any data set encoded in MIDI format, or any other format containing pitch, onset and duration information for each note in a melody.

### 2.2 Gestalt-based segmentation

Segmentation is a core activity for musical processing and cognition (Lerdahl & Jackendoff, 1983). In order to study this mechanism, some authors adapt concepts of visual processing to study musical processing. Cambouropoulos (1997, 2001) presents a segmentation model based on Gestalt principles of similarity and proximity, known as the local boundary detection model (LBDM). The LBDM computes a profile of segmentation strength in the range [0, 1], based pitch intervals, inter-onset-intervals and rests. When the strength exceeds a threshold, a segmentation point is introduced. (Cambouropoulos, 2001). We use the LBDM here as a baseline for our model.

### 2.3 The use of wavelets in the symbolic domain

Wavelet analysis has been applied to diverse time series datasets. A time series is a set of observations recorded at a specified time (Brockwell & Davis, 2009). The use of wavelets for time series processing and analysis can be found in different areas, i.e. meteorological (Torrence & Compo, 1998), political (Aguiar-Conraria, Magalhaes, Soares, 2012), medical (Hsu, 2010), financial (Hsieh, Hsiao, & Yeh, 2011). Wavelets are also well known in audio music information retrieval (Andén & Mallat, 2011; Jeon & Ma, 2011; Smith & Honing, 2008; Tzanetakis, Essl, & Cook, 2001), but they have been scarcely applied on symbolic music representations. The only example of wavelets applied to symbolic music representation, apart from our previous study (Velarde & Weyde, 2012), is presented by Pinto (2009), demonstrating that it is possible to index melodic sequences with few wavelet coefficients, obtaining improved retrieval results compared to the direct use of melodic sequences. The method used by Pinto can be exploited for compression purposes, whereas our method is used for structural analysis and classification.

## 3. THE METHOD

We extend the method introduced in Velarde and Weyde (2012) by exploring segmentation based on the information of the wavelet coefficients' local maxima, and evaluate it on the classification of folk tunes into tune families. Our previous study (Velarde & Weyde, 2012) showed good results in a different classification task using the 15 Two-Part Inventions by J. S. Bach.

### 3.1 Representation

We represent melodies as normalized pitch signals or by the wavelet coefficients of the pitch signals. Discrete pitch signals $v[l]$ with length $L$ are sampled from MIDI files at a rate $r$ (given in number of samples per quarter note), so that we have a pitch value for every time point, expressed as $v[t]$. Rests are replaced by the following procedure: if a rest occurs at the beginning of a sequence, it is replaced by the first pitch number that appears in the sequence, otherwise it is replaced by the pitch number of the last note that precedes it.

**Normalized pitch signal representation (vr).** We normalize pitch signals segments, by subtracting the average pitch in order to make the representation transposition-invariant. The normalization is applied after the segmentation.

**Wavelet representation (wr).** We apply the continuous wavelet transform (CWT) (Mallat, 2009), expressed in a discretized version as the inner product of the pitch signal $v[l]$ and the Haar wavelet $\psi_{s,u}[l]$, at position $u$ and scale $s$:

$$w_s[u] = \sum_{l=1}^{L} \psi_{s,u}[l]v[l] \qquad (1)$$

To avoid edge effects due to finite-length sequences (Torrence & Compo, 1998), we pad on both ends with a mirror image of the pitch signal (Woody & Brown, 2007). Once the coefficients are obtained, the segment that corresponds to the padding is removed, so that the signal maintains its original length.

### 3.2 Segmentation

**Wavelet segmentation (ws).** Local maxima of the wavelet coefficients occur when the inner product of the melody and the wavelet is maximal. This occurs with the Haar wavelet, when there is a locally maximal change of pitch (averaged over half the length of the wavelet) in the melody. We use local maxima of wavelet coefficients to determine local boundaries.

### 3.3 Classification

The melodic segments are used as the data points for classification. A melody is represented as a set of segments, and we use the *k*-Nearest-Neighbour (kNN) method for classification (Mitchell, 1997). We use two different distance measures: cityblock distance and Euclidean distance. We define the maximal length n of all segments to be compared and pad shorter segments as necessary with zeros at the end.

### 4. EXPERIMENT

In our experiment we address the question of how filtering the representation of melodic segments affects the folk tune family classification. We assumed that if segments represent meaningful melodic structures, they can be used to identify tunes belonging to a tune family and that some time-scales of the melodic contour might be more discriminative than others.

We ran the experiment[1] using the collection "Onder de groene linde" (Grijp, 2008). This collection is a high quality data set of 360 monophonic songs classified into 26 families according to field-experts' similarity assessments in terms of melodic, rhythmic and motivic content (Volk & van Kranenburg, 2012). The MIDI files of this collection are sampled into pitch signals with a sampling rate of 8 samples per quarter note (qn). We apply the CWT with the Haar wavelet using a dyadic set of 8 scales. Melodies are represented as normalized pitch signals (vr) or as the resulting wavelet coefficients (wr). Signals are segmented by the wavelet coefficients' local maxima (ws), or by the local boundary detection model (LBDM; Cambouropoulos, 1997, 2001) using thresholds from 0.1 to 0.8 in steps of 0.1. We explored the parameter space with a grid search testing all combinations of representations and segmentations: wavelet representation (wr), normalized pitch signal representation (vr), wavelet segmentation (ws), LBDM (LBDM) segmentation and 1 to 5 nearest neighbours. Segments are used to build classifiers from training sets and that are tested on unseen folk melodies. We evaluate the classification accuracy with cityblock and Euclidean distances in leave-one-out cross validation.

### 5. RESULTS

The results of the experiment can be seen in Figures 1 to 4. Alternatively, Tables 1 and 2 shows the best and worst classification values over all parameters for each combination of representation-segmentation, for each value of *k* in the kNN method, and for Euclidean and cityblock distance metrics. The results show that wavelet filtering of the melodies can improve classification performance compared to using the pitch signal directly. Independently of the segmentation method, wavelet representation proves to be more discriminative than pitch signals. For this corpus and experimental setup, we have used single time-scales and evaluated this melodic discrimination performance. The classification performance varies, obtaining best results at small scales and poor results at large scales, with exception of the largest scale which recovers its performance to some extent.

In terms of segmentation, it is possible to observe that shorter segments produce better results when used with wavelet representation. This is contrary to the results of the LBDM applied to pitch signals, where shorter segments produce worse results than larger ones. We observe an improvement towards threshold 0.4 and a gradual improvement towards the threshold of 0.8, which corresponds to larger segments, meaning that using the complete melodic sequences or a combination of complete melodies and melodic segments, can lead to better classification results when using pitch signals.

---

[1] The algorithms are implemented in MATLAB (The Mathworks, Inc) using the Wavelet Toolbox and the MIDI Toolbox for the implementation of the LBDM (Eerola & Toiviainen, 2004), and we use an update of Christine Smit's read_midi function (http://www.ee.columbia.edu/~csmit/matlab_midi.html, accessed 4 October 2012).

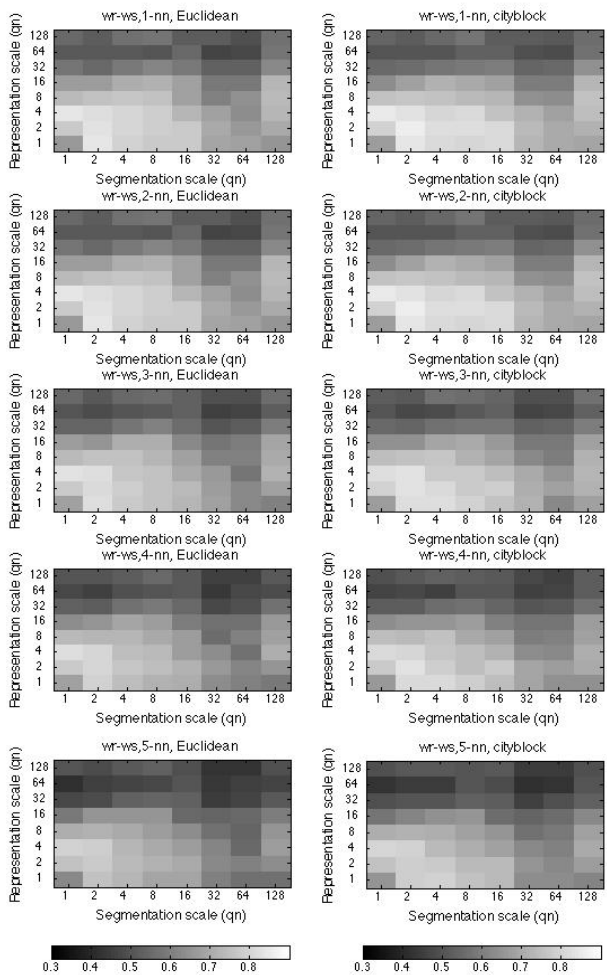

**Figure 1.** Accuracies for the combination of wavelet representation (wr) and wavelet segmentation (ws).

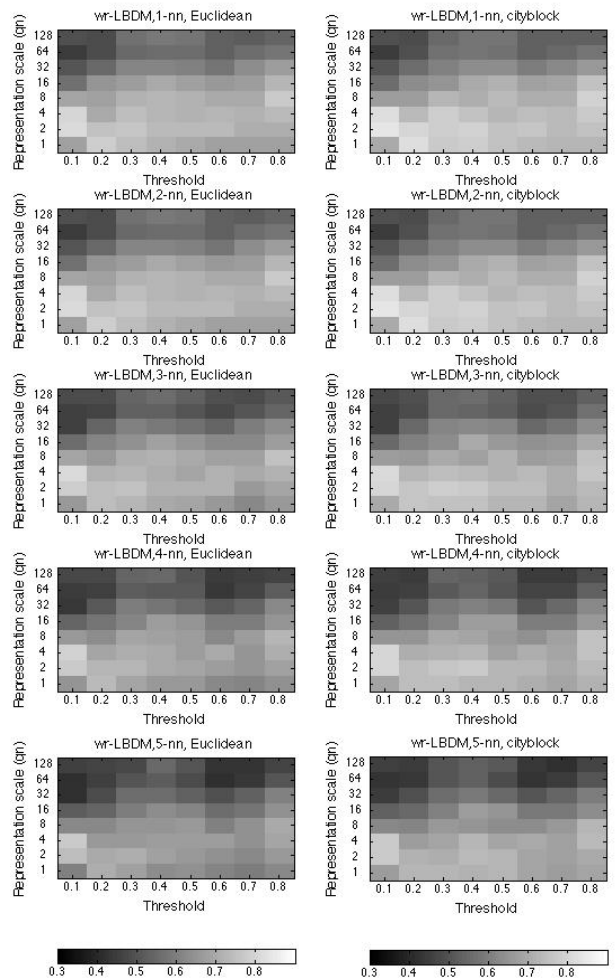

**Figure 2.** Accuracies for the combination of wavelet representation (wr) and local boundary detection model (LBDM).

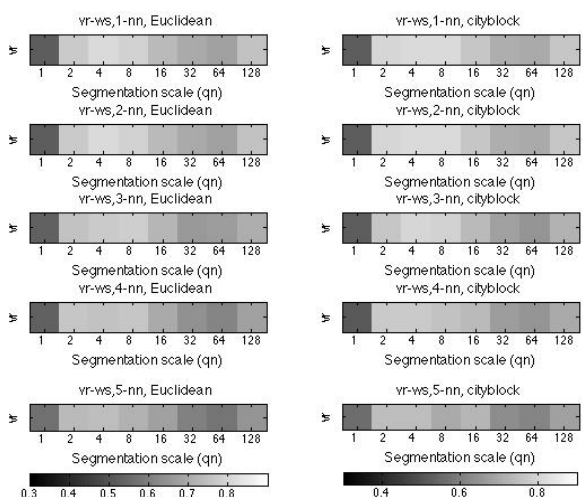

**Figure 3.** Accuracies for the combination of pitch signal representation (vr) and wavelet segmentation (ws).

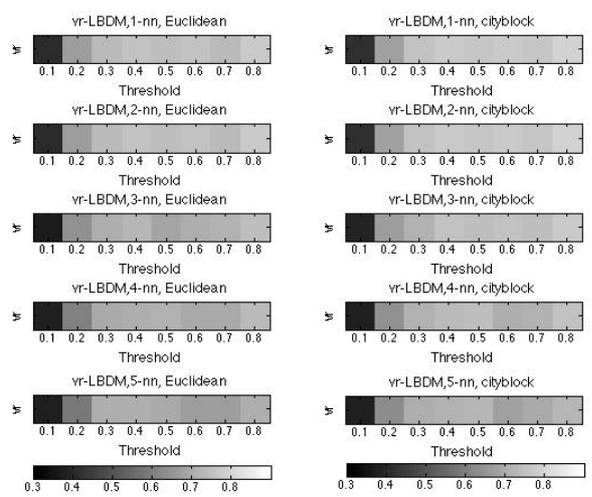

**Figure 4**. Accuracies for the combination of pitch signal representation (vr) and local boundary detection model (LBDM).

In general, similarity measured by cityblock distance proves more accurate than by Euclidean distance in pitch signals over time or wavelet representations, and the effect of using cityblock distance makes the difference between segmentation methods less important. The number of *k*-nearest neighbours shows that one or two neighbours produce the best results and when *k* increases further the accuracy decreases.

| represent.-segment. | Value | Euclidean distance |||||
|---|---|---|---|---|---|---|
| | | Nearest Neighbours |||||
| | | 1 | 2 | 3 | 4 | 5 |
| wr-ws | best | **0.8417** | **0.8417** | 0.8306 | 0.8194 | 0.7917 |
| | worst | 0.4667 | 0.4667 | 0.4583 | 0.4333 | 0.4167 |
| wr-LBDM | best | 0.8111 | 0.8111 | 0.8083 | 0.7889 | 0.7694 |
| | worst | 0.4472 | 0.4472 | 0.4528 | 0.4333 | 0.4139 |
| vr-ws | best | 0.8083 | 0.8083 | 0.7806 | 0.7667 | 0.7444 |
| | worst | 0.5194 | 0.5194 | 0.5333 | 0.525 | 0.5639 |
| vr-LBDM | best | 0.7778 | 0.7778 | 0.7444 | 0.7333 | 0.7083 |
| | worst | 0.4111 | 0.4111 | 0.3722 | 0.3806 | 0.3806 |

**Table 1.** Classification accuracies best and worst values for each combinations using Euclidean distance.

| represent.-segment. | Value | Cityblock distance |||||
|---|---|---|---|---|---|---|
| | | Nearest Neighbours |||||
| | | 1 | 2 | 3 | 4 | 5 |
| wr-ws | best | **0.8556** | **0.8556** | 0.8333 | 0.8306 | 0.7972 |
| | worst | 0.4833 | 0.4833 | 0.4639 | 0.45 | 0.4167 |
| wr-LBDM | best | 0.8417 | 0.8417 | 0.8083 | 0.8028 | 0.7778 |
| | worst | 0.4417 | 0.4417 | 0.4556 | 0.4417 | 0.4139 |
| vr-ws | best | 0.8139 | 0.8139 | 0.7972 | 0.7778 | 0.7472 |
| | worst | 0.5194 | 0.5194 | 0.5194 | 0.5139 | 0.5583 |
| vr-LBDM | best | 0.7889 | 0.7889 | 0.7778 | 0.75 | 0.725 |
| | worst | 0.4139 | 0.4139 | 0.3861 | 0.3778 | 0.3806 |

**Table 2.** Classification accuracies best and worst values for each combinations using cityblock distance.

## 6. DISCUSSION AND FUTURE DIRECTIONS

The best classification accuracies based on wavelet segmentation are only slightly better than the best accuracies obtained by the LBDM. The parameter exploration shows however, that wavelet segmentation performs better across different scales than the LBDM across different thresholds. Interestingly, these comparable methods meet the criteria of measuring local changes in melodic contour. While the LBDM measures the degree of change between successive values, the wavelet segmentation finds locally maximal falls of average pitch in melodies using different scales. The fact that small scales perform better than larger scales corroborates the findings of van Kranenburg et al. (2013) that local processing is most important in melodic similarity.

In terms of representation, wavelet-representation proves more discriminative than raw pitch signals. We assume that this is due to the transposition invariance of the wavelet representation and the emphasis on a specific time-scale.

Our best results are far less accurate than the results reported by van Kranenburg et al. (2013) using alignment methods on the same corpus. Our method uses only the information that is encoded in MIDI format (pitch number, onset and duration). It requires less encoded expert knowledge than the method used by van Kranenburg (2010), making it applicable to other corpuses of folk songs encoded in MIDI format or similar. In order to make a more reliable comparison, our method would need to include the expert based features used by van Kranenburg (2010). For instance, annotated phrase information seems to improve importantly the results obtained by sequence alignment algorithms. This information could be used to improve the scale selection. Also, our method uses only the information about contained segments, and not the order of the segments, leaving room for further work.

We used one default setup for the whole corpus, i.e. one best performing scale for all songs. In a future study, we are interested to address wavelet scale selection derived from individual songs' periodicities.

## 7. CONCLUSION

The main contribution of this research is the evaluation of wavelet-filtered signals for melodic segmentation and classification on a corpus of folk songs in MIDI format. Wavelet-filtering proves more discriminative than direct representation of pitch signals or pitch-time series. Segmentation by local maxima of wavelet coefficients performs slightly better than LBDM segmentation when processing at individual scales. Small scales perform better than large scales, indicating that local processing may be more relevant for melodic similarity in classification tasks.

The method presented here can be applied to other corpora and other symbolic formats that encode melodies. Possible ways to improve the classification performance of the method presented in this paper could be using alignment of wavelet representations of complete melodies, using selective combination of scales and exploring metrical information derived from songs' periodicities.

**Acknowledgements**

We thank Peter van Kranenburg (Meertens Institute, Amsterdam) for sharing the Dutch Tune Families data set. Gissel Velarde is supported by the Department of Architecture, Design and Media Technology at Aalborg University.

## 8. REFERENCES

Andén, J., & Mallat, S. (2011). Multiscale scattering for audio classification. In: *Proceedings of the 12th International Society for Music Information Retrieval Conference (ISMIR 2011), Utrecht, NL.: ISMIR*, pp. 657–662. Available online at http://ismir2011.ismir.net/papers/PS6-1.pdf

Aguiar-Conraria, L., Magalhães, P. C. and Soares, M. J. (2012), Cycles in politics: wavelet analysis of political time series. *American Journal of Political Science*, 56: 500–518. doi: 10.1111/j.1540-5907.2011.00566.x

Brockwell, P. & Davis, R. (2009). Time series: theory and methods. Springer series in statistics. Second edition.

Cambouropoulos, E. (1997). Musical rhythm: a formal model for determining local boundaries, accents and metre in a melodic surface. In: *M. Leman (Ed.), Music, Gestalt and Computing: Studies in Cognitive and Systematic Musicology*. Berlin: Springer, pp. 277-293.

Cambouropoulos, E. (2001). The local boundary detection model (LBDM) and its application in the study of expressive timing. In: *Proceedings of the International Computer Music Conference. San Francisco, CA: ICMA*, pp. 17-22.

Cambouropoulos, E. (2009). How similar is similar?. *Musicæ Scientiæ*. Discussion Forum 4B, pp. 7-24

Eerola, T. & Toiviainen, P. (2004). MIDI Toolbox: MATLAB Tools for Music Research. University of Jyväskylä. Available at http://www.jyu.fi/hum/laitokset/musiikki/en/research/coe/materials/miditoolbox/.

Grijp, L.P.. (2008). Introduction. In: L.P. Grijp & I. van Beersum (Eds.), Onder de groene linde. 163 verhalende liederen uit de mondelinge overlevering, opgenomen door Ate Doornbosch e.a./Under the green linden. 163 Dutch Ballads from the oral tradition recorded by Ate Doornbosch a.o. (Boek + 9 cd's + 1 dvd). Amsterdam/Hilversum : Meertens Instituut & Music and Words. pp. 18-27.

Hillewaere, R, Manderick, B., & Conklin, D. (2009). Global feature versus event models for folk song classification. In: *10th International Society for Music Information Retrieval Conference (ISMIR 2009), Kobe, Japan*. pp. 729-733. Available online at http://ismir2009.ismir.net/proceedings/OS9-1.pdf.

Hillewaere, R., Manderick, B. and Conklin, D. (2012) String methods for folk music classification. In: *13th International Society for Music Information Retrieval Conference (ISMIR 2012)*.

Huron, D. (1996) The melodic arch in western folksongs. *Computing in Musicology*, Vol. 10, pp. 3-23.

Hsieh, TJ. Hsiao, H.f., & Yeh, WC. (2011). Forecasting stock markets using wavelet transforms and recurrent neural networks: an integrated system based on artificial bee colony algorithm. *Applied soft computing*, Vol. 11 Issue 2. March 2011, pp. 2510–2525. Elsevier

Hsu, WY. (2010). EEG-based motor imagery classification using neuro-fuzzy prediction and wavelet fractal features. *J. Neurosci Methods*. 2010 Jun 15;189(2):295-302. doi: 10.1016/j.jneumeth.2010.03.030. Epub 2010 Apr 8.

Jeon, W., & Ma, C. (2011). Efficient search of music pitch contours using wavelet transforms and segmented dynamic time warping. In: *Proceedings of the IEEE International Conference on Acoustics, Speech and Signal Processing (ICASSP 2011)*, pp. 2304–2307.

Lerdahl, F., & Jackendoff, R. (1983). A Generative Theory of Tonal Music.Cambridge, MA.: MIT Press.

Mallat, S. (2009). A wavelet tour of signal processing - The sparse way. Academic Press, Third Edition, 2009.

Mitchell, T. (1997). Machine Learning, (McGraw-Hill).

Müllensiefen, D., Bonometti, M., Stewart, L., and Wiggins, G. (2009). Testing Different Models of Melodic Contour. *7th Triennial Conference of the European Society of the Cognitive Sciences of Music (ESCOM 2009), University of Jyäskylä, Finalnd.*

Müllensiefen, D, and Wiggins, G. (2011). Polynomial functions as a representation of melodic phrase contour. In *A. Schneider & A. von Ruschkowski (Eds.), Systematic Musicology: Empirical and Theoretical Studies.* pp. 63-88. Frankfurt a.M.: Peter Lang.

Pinto, A. (2009). Indexing melodic sequences via wavelet transform. In: *Proceedings of the IEEE International Conference on Multimedia and Expo (ICME'09)*. pp. 882–885.

Smith, L.M., & Honing, H. (2008). Time-frequency representation of musical rhythm by continuous wavelets. *Journal of Mathematics and Music*, Vol. 2, No. 2, pp. 81–97.

Steinbeck, W. (1982). Struktur und Ähnlichkeit: Methoden automatisierter Melodieanalyse. Kassel: Bärenreiter.

Tzanetakis, G., Essl, G., & Cook, P. (2001). Audio analysis using the discrete wavelet transform. In: *Proc. WSES Int. Conf. Acoustics and Music: Theory and Applications (AMTA 2001), Skiathos, Greece*. Available online at http://webhome.cs.uvic.ca/~gtzan/work/pubs/amta01gtzan.pdf

van Kranenburg, P. (2010) A Computational Approach to Content-Based Retrieval of Folk Song Melodies. [S.l.] : [s.n.], 2010. Full text: http://depot.knaw.nl/8400

van Kranenburg, P., Garbers, J., Volk, A., Wiering, F. Grijp, L.P. and Veltkamp, R. (2009). Collaboration perspectives for folk Song research and music information retrieval: The indispensable role of computational musicology, *Journal of Interdisciplinary Music Studies*. (2009), doi: 10.4407/jims.2009.12.030

van Kranenburg, P., Volk, A., & Wiering, F. (2013): A Comparison between Global and Local Features for Computational Classification of Folk Song Melodies, *Journal of New Music Research*, 42:1, 1-18

Velarde, G., & Weyde, T. (2012). On symbolic music classification using wavelet transform. In: *International Conference on Applied and Theoretical Information Systems Research, Taipei, Taiwan*.

Volk, A. & van Kranenburg, P. (2012). Melodic similarity among folk songs: An annotation study on similarity-based categorization in music. *Musicae Scientiae*, November 2012 vol. 16 no. 3, pp. 317-339.

Wiering, F., Veltkamp, R., Garbers, J., Volk, A. and van Kranenburg, P. (2009). Modelling Folksong Melodies. *Interdiciplinary Science Reviews*, Vol 34, No. 2-3, 154-171.

Woody, N. A. & Brown, S. D. (2007) Selecting wavelet transform scales for multivariate classification. *J. Chemometrics*, 21: 357–363. doi: 10.1002/cem.1060